\documentclass{article}
\usepackage{spconf,amsmath,graphicx,hyperref}
\usepackage{cite}
\usepackage{amsfonts}
\usepackage{algorithmic}
\usepackage{graphicx}
\usepackage{textcomp}
\usepackage{booktabs} 
\usepackage{enumerate}
\usepackage{caption}
\usepackage{enumitem}
\usepackage{multirow}
\usepackage{subcaption}
\usepackage{colortbl}  
\usepackage{xcolor}
\usepackage{color}
\definecolor{Gray}{gray}{0.94}

\title{WebRouter: Query-specific Router via Variational Information
Bottleneck for Cost-sensitive Web Agent}
\name{Tao Li\textsuperscript{1}, \quad Jinlong Hu\textsuperscript{2}, \quad  Yang Wang\textsuperscript{3}, \quad Junfeng Liu\textsuperscript{4$^\dagger$},   \quad Xuejun Liu\textsuperscript{1} \thanks{$^\dagger$Corresponding author} 
\thanks{This paper is under review at ICASSP 2026.} 
}
\address{\textsuperscript{1}College of Artificial Intelligence, Nanjing University of Aeronautics and Astronautics, Nanjing, China\\
        \textsuperscript{2}Hong Kong Baptist University, Hong Kong SAR \\
        \textsuperscript{3}SKLCCSE Lab, Beihang University, Beijing, China\\
        \textsuperscript{4}Pengcheng Laboratory, Shenzhen, China\\
        }
\begin{document}

\maketitle
\begin{abstract}
LLM-brained web agents offer powerful capabilities for web automation but face a critical cost-performance trade-off. 
The challenge is amplified by web agents' inherently complex prompts that include goals, action histories, and environmental states, leading to degraded LLM ensemble performance.
To address this, we introduce WebRouter, a novel query-specific router trained from an information-theoretic perspective. Our core contribution is a cost-aware Variational Information Bottleneck (ca-VIB) objective, which learns a compressed representation of the input prompt while explicitly penalizing the expected operational cost. Experiments on five real-world websites from the WebVoyager benchmark show that WebRouter reduces operational costs by a striking 87.8\% compared to a GPT-4o baseline, while incurring only a 3.8\% accuracy drop.
\end{abstract}

 
  
\section{Introduction}
\label{sec:intro}

The advent of Large Language Models (LLMs) has catalyzed a paradigm shift in the automation of GUI, moving from traditional script-based or rule-based methods to a new era of LLM-brained autonomous agents\cite{mind2web, WebDancer, webwalker, webarena}. 
These agents understand instructions and navigate complex digital environments with remarkable flexibility. Web agents, in particular, stand out for their practical uses, such as automated information retrieval, e-commerce navigation, and data extraction, highlighting their important value in research and industry~\cite{browseruse, webagent1,  webagent3, guisurvey2}.

Modern web agents are fundamentally LLM-brained. With substantial research dedicated to refining their components, \emph{e.g.,} planners and perception modules, an agent's ability to understand its environment has significantly improved~\cite{guisurvey2}. 
As a result, its task-completion capability now hinges directly on the reasoning prowess of its core LLM.

However, employing state-of-the-art LLMs presents a challenging cost-performance trade-off~\cite{browseruse, mind2web}. Upgrading to a more powerful model often yields only marginal accuracy gains while substantially increasing operational costs, \emph{e.g.,} price. 
This trade-off makes it impractical to use a single ``best model'' in real-world web agents.
Therefore, it is important to develop a dynamic router that matches each web query to the most cost-effective LLM capable of resolving it.
\begin{figure}[tb!]
    \centering
    \includegraphics[width=0.384\textwidth]{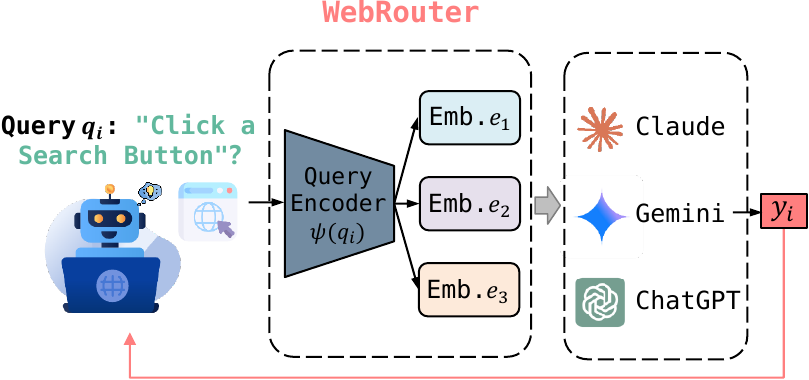}%
    \caption{The inference pipeline of WebRouter.}
    \label{fig:webrouter}
\end{figure} 

\begin{figure}[tb!]
    \centering
    \includegraphics[width=0.22\textwidth]{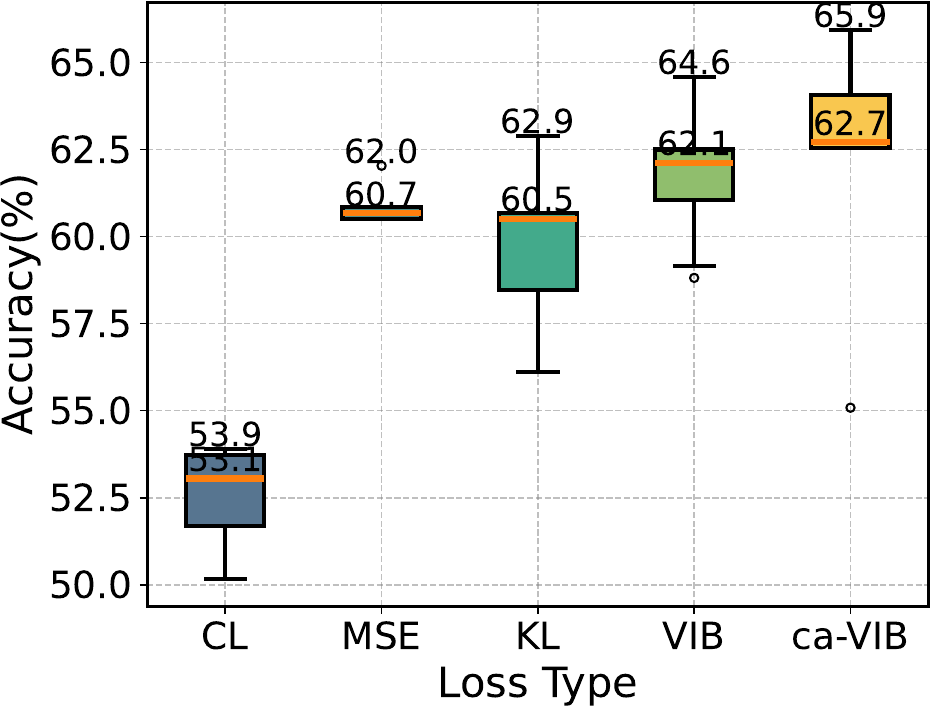}%
    \caption{Query routing accuracy with different loss functions.} %
    \label{fig:loss}
\end{figure}

Indeed, several pioneering works have explored the concept of LLM ensemble and routing~\cite{ensemble1, ensemblesurvey1, cascading, routerdc, zooter}.
Within the routing sub-domain, for example, ZOOTER employs knowledge distillation from a reward model using KL divergence\cite{zooter}, while RouterDC utilizes contrastive learning to differentiate between strong and weak models for a given query\cite{routerdc}.
However, these methods face significant challenges in web agent scenarios due to verbose and complex prompts, \emph{i.e.,} concatenating user goals, action histories (memory), and detailed environmental states (perception). This informational redundancy can degrade the performance of conventional routing mechanisms, as illustrated in Fig.~\ref{fig:loss}

To address this, we propose a router trained under the Variational Information Bottleneck (VIB) principle~\cite{vib}, as compared in Fig.~\ref{fig:webrouter}. VIB offers a principled framework to learn a compressed representation of the input, explicitly filtering out irrelevant information while preserving only the features critical for the routing decision. This approach yields a more robust and efficient routing function, specifically tailored for the noisy, high-dimensional inputs of web agents~\cite{mind2web}.

Our contributions are threefold: 1) We are the first to formulate web agent routing from an information-theoretic perspective, using VIB to handle the noisy and redundant prompts inherent to this domain.
2) We design a novel cost-aware VIB objective that integrates a pre-defined cost function, creating a principled trade-off between accuracy and cost.
3) We achieve state-of-the-art cost-efficiency on five real-life websites~\cite{Webvoyager}, reducing operational costs by 87.8\% with only a 3.8\% drop in accuracy.

\section{Related Work}
\label{sec:related}

\noindent\textbf{GUI and Web Agents.}
LLM-brained GUI agents represent a new paradigm, using an LLM as a cognitive ``brain'' to operate graphical interfaces based on natural language commands \cite{guisurvey2,browseruse}. These agents typically operate in a loop, perceiving the environment via screenshots or DOM trees, maintaining an action history in memory, and planning subsequent actions~\cite{webarena, WebDancer, webagent1, webagent2, mind2web}. Current research focuses on improving agent autonomy and reliability when navigating the complex nature of real-world web environments~\cite{uiagent1, uiagent2}.

\noindent\textbf{LLM Ensembling and Routing.}
To harness the diverse capabilities of multiple LLMs, ensembling strategies such as token-level~\cite{ensembletoken} and span-level~\cite{ensembletspan}, are employed~\cite{ensemblesurvey1, ensemble1, router1}. 
LLM routing and cascading have thus emerged as a cost-effective alternative, training a lightweight model to direct each query to the most suitable LLM from a pool of candidates~\cite{router1, yue2023mixture, cascading} \emph{e.g.,}  ZOOTER~\cite{zooter} and RouterDC~\cite{routerdc}.

\noindent\textbf{Variational Information Bottleneck (VIB).}
The Information Bottleneck (IB) principle provides a theoretical framework for learning a compressed representation $Z$ of an input $X$~\cite{vib2000}. 
This is formally expressed by the objective:
\begin{equation}
  \mathcal{L}_{\text{IB}} =  I(Z; Y) - \beta I(Z; X),
  \label{eq:ib}
\end{equation}
which seeks to maximize the mutual information with the target $Y$  while penalizing the mutual information with the input $X$, balanced by the hyperparameter $\beta$. 
As direct computation of mutual information is often intractable in deep learning, the VIB was developed, which minimizes a tractable variational bound of this objective~\cite{vib, vib2018, vib2019}.

\section{Method of WebRouter}
\label{sec:method}

\noindent\textbf{Problem Formulation.}
Let $\mathcal{M} = \{\mathcal{M}_t\}_{t=1}^T$ be a set of candidate LLMs. In the context of web agents, a query $\boldsymbol{q}_i$ is a complex prompt with the user's high-level goal, the agent's action history, and the current web representation. Given a training dataset $\mathcal{D}_{\text{train}} = \{(\boldsymbol{q}_i, y_i)\}_{i=1}^n$, where $y_i$ is the ground-truth outcome, our objective is to learn a router $\psi$. The router $\psi$ takes a query $\boldsymbol{q}_i$ as input and outputs a probability distribution $\boldsymbol{p}_i = \psi(\boldsymbol{q}_i) \in \mathbb{R}^T$ over the set of LLMs, indicating the suitability of each model for the given task.

\subsection{Scoring}  
To train the router, we require a supervision signal that balances task performance with operational cost. We design a scoring function that assigns a score $s_i^{(t)}$ to each query-model pair $(\boldsymbol{q}_i, \mathcal{M}_t)$. 
Our scoring is conditioned on task-level success: we only generate scores for the constituent queries $\boldsymbol{q}_i$ of a task if the chosen model $\mathcal{M}_t$ successfully completes the overall task. This score is only non-zero if the model is successful, and its magnitude is inversely proportional to the operational cost, thus rewarding both correctness and cost.

Specifically, we first define the operational cost $C(\boldsymbol{q}_i, \mathcal{M}_t)$ for model $\mathcal{M}_t$ to process query $\boldsymbol{q}_i$:
\begin{equation}
\label{eq:cost}
C(\boldsymbol{q}_i, \mathcal{M}_t) = n_p \cdot c_p^{(t)} + n_c \cdot c_c^{(t)},
\end{equation}
where $n_p$ and $n_c$ are the number of prompt and completion tokens, respectively, and $c_p^{(t)}$ and $c_c^{(t)}$ are the corresponding per-token costs for model $\mathcal{M}_t$.

Next, let $P(\boldsymbol{q}_i, \mathcal{M}_t) \in \{0, 1\}$ be a binary indicator of task success, where $1$ denotes a correct outcome.  To transform the unbounded operational costs into a bounded utility measure, we employ an exponential utility function, $U(c) = \exp(-c)$. This choice is motivated by its property of disproportionately penalizing higher costs, reflecting the severe impact of expensive LLM calls. This utility is then min-max normalized across all models to produce the final cost score, $S_{\text{cost}}(C_i^{(t)})$, which scales the utility to a range in $\{0, 1\}$ where 1 represents the most cost-effective model.
  
The final training score $s_i^{(t)}$ is the product of task success and this normalized cost score, ensuring that only correct and cost-effective models receive a positive supervision signal:
\begin{equation}
\label{eq:final_score}
s_i^{(t)} = P(\boldsymbol{q}_i, \mathcal{M}_t) \times S_{\text{cost}}(C_i^{(t)}).
\end{equation}
This score creates a high-contrast supervision signal that clearly distinguishes the most cost-effective models from both the unsuccessful and the inefficient ones, providing a clear signal for training our VIB-based router.

\subsection{Cost-aware VIB Loss}
\begin{figure}[t]
    \centering
    \includegraphics[width=0.50\columnwidth]{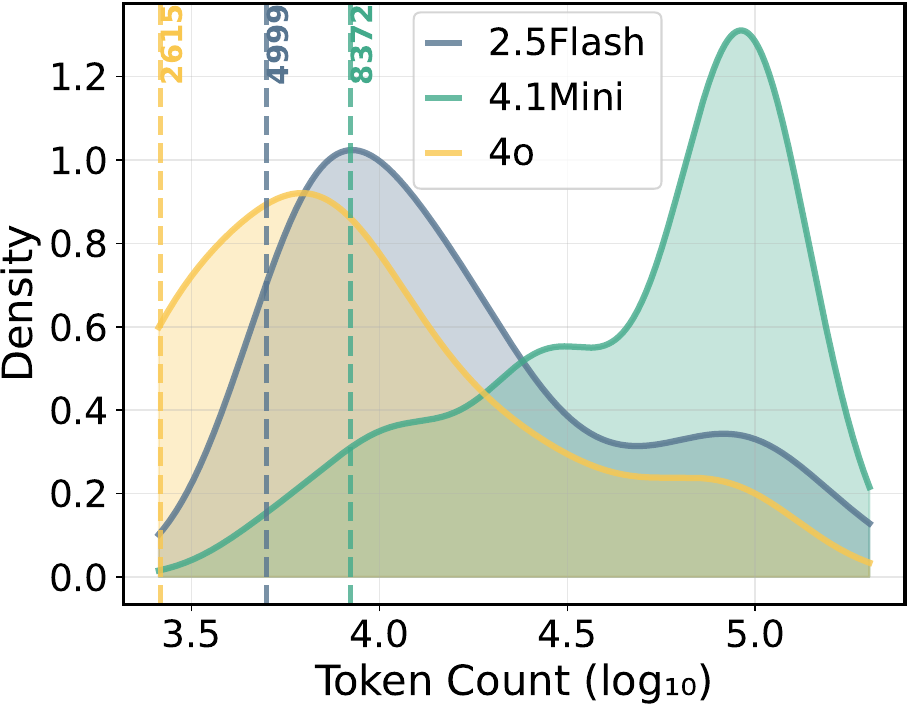}
    \caption{Distribution of query token lengths for various LLMs.}
    \label{fig:token_dis}
\end{figure}

As a web agent progresses through a task, its input prompts become increasingly verbose and noisy. 
This is because each query, $\boldsymbol{q}_i$, is a dynamic concatenation of the user's high-level goal, the current web representation, and a \emph{growing action history} from the agent's memory. 
As illustrated in Fig.~\ref{fig:token_dis}, this accumulation process results in a significant portion of queries exceeding thousands of tokens, leading to substantial informational redundancy. 
We posit that this inherent noise challenges traditional routing mechanisms that rely on direct semantic embeddings~\cite{zooter, router1, routerdc}. 
Fortunately, the Information Bottleneck principle \cite{vib2000} directly addresses this challenge.
It provides a formal framework, defined in Eq.~\eqref{eq:ib}, for learning a compressed representation of the input that is minimally sufficient for the downstream task, thus inherently filtering out superfluous information.

To incorporate both cost and performance into a unified objective, we derive a novel cost-aware VIB loss. We begin with the standard variational bound of the IB objective~\cite{vib}, which is minimized during training:
\begin{equation}
\label{eq:vib_standard}
\mathcal{L}_{\text{VIB}} = \mathbb{E}_{p(z|\boldsymbol{q})} [-\log p_\phi(y|z)] + \beta \text{KL}[p_\theta(z|\boldsymbol{q}) || r(z)],
\end{equation}
where $p_\theta(z|\boldsymbol{q})$ is the stochastic encoder that maps an input query $\boldsymbol{q}$ to a latent representation $z$, and $p_\phi(y|z)$ is the decoder that predicts the target model distribution $y$ from $z$. The first term is a standard prediction loss (cross-entropy) ensuring accuracy, while the second KL-divergence term enforces the compression

To specifically address the token-level redundancy in web agent prompts, we implement the compression term by learning a stochastic binary mask, inspired by~\cite{vib2020ration}. We define the latent representation $z$ as the element-wise product of the input's feature representation $\boldsymbol{h}_{\boldsymbol{q}}$ and a stochastic binary mask $\boldsymbol{m}$: $z = \boldsymbol{m} \odot \boldsymbol{h}_{\boldsymbol{q}}$. 
With this masking formulation, the KL-divergence term simplifies, as it becomes proportional to the KL-divergence of the mask distribution after dropping terms that are constant \emph{w.r.t} the model parameters:
\begin{equation}
\label{eq:kl_decomposition}
\text{KL}(p_\theta(z|\boldsymbol{q}) || r(z)) \propto \text{KL}(p_\theta(\boldsymbol{m}|\boldsymbol{q}) || r(\boldsymbol{m})).
\end{equation}
After dropping the entropy of the input features, which is constant \emph{w.r.t.} the model parameters, the effective regularization term becomes simply $\text{KL}(p_\theta(\boldsymbol{m}|\boldsymbol{q}) || r(\boldsymbol{m}))$.

On this basis, we introduce our cost-aware VIB (ca-VIB) loss. We augment the standard VIB objective with an explicit cost regularization term that represents the expected operational cost of the router's decision. For a training set $\mathcal{D}_{\text{train}}$ of $N$ samples, our final objective to minimize is:
\begin{equation}
\begin{aligned}
\label{eq:ca_vib_loss}
\mathcal{L}_{\text{ca-VIB}} = \frac{1}{N}\sum_{i=1}^N \Big( & \mathbb{E}_{p_\theta(z_i|\boldsymbol{q}_i)} [-\log p_\phi(y_i|z_i)] \\
& + \beta \cdot \text{KL}[p_\theta(\boldsymbol{m}_i|\boldsymbol{q}_i) || r(\boldsymbol{m}_i)] \\
& + \lambda \cdot \mathbb{E}_{p_\theta(z_i|\boldsymbol{q}_i)} \left[ \sum_{t=1}^T p_\phi(y_t|z_i) \cdot C(\mathcal{M}_t) \right] \Big),
\end{aligned}
\end{equation}
where $\lambda$ is a hyperparameter controlling cost-sensitivity, and $C(\mathcal{M}_t)$ is the pre-defined, query-agnostic unit cost of invoking model $\mathcal{M}_t$. We use a unit cost (e.g., $c_p^{(t)} + c_c^{(t)}$ from Eq.~\eqref{eq:cost} for a single token) rather than the full query-specific cost because the latter depends on the completion length, which is unknown at routing time. This formulation allows the loss to directly penalize the inherent expense of selecting a model.

The learning process is guided by the training scores $s_i^{(t)}$ defined in Eq.~\eqref{eq:final_score}. The prediction loss term, $\mathbb{E}[-\log p_\phi(y_i|z_i)]$, is a cross-entropy loss between the router's output and a target distribution derived from a softmax over the score vector $\boldsymbol{s}_i$. By jointly minimizing this prediction error and the expected operational cost, WebRouter learns to generate a compressed representation $z_i$ that is sufficient to select the most cost-effective LLM, effectively learning to replicate the high-contrast signal embedded in our scoring function.

\begin{figure}[!tb]
    \centering 
    \includegraphics[width=0.50\columnwidth]{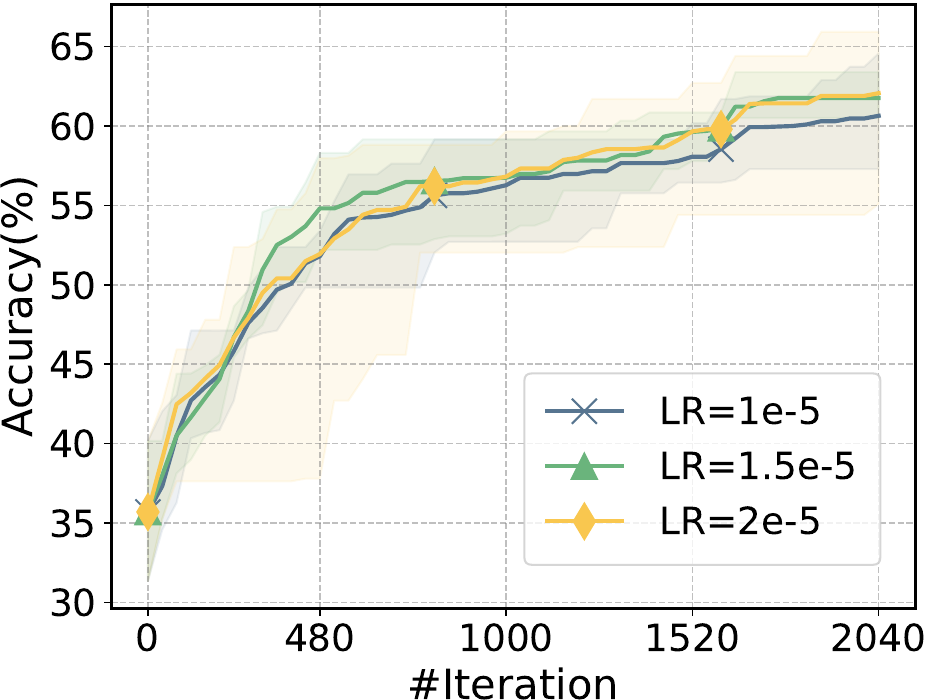}%
    \caption{Training loss curves for different learning rates.}
    \label{fig:train}
\end{figure}

\begin{table*}[tb!]
\renewcommand{\arraystretch}{1.1}
\centering
\caption{Main Results of WebRouter. The best is in \textbf{bold} and the second-best is \underline{underlined}.} 
\label{tab:main_res}
\setlength{\tabcolsep}{4pt} 
\resizebox{\textwidth}{!}{
\begin{tabular}{c|ccc|ccc|ccc|ccc|ccc} \toprule
\multirow{2}{*}{Dataset} & \multicolumn{3}{c|}{2.5Flash~\cite{browseruse}} & \multicolumn{3}{c|}{4.1Mini~\cite{browseruse}} & \multicolumn{3}{c|}{4o~\cite{browseruse}} & \multicolumn{3}{c|}{routerDC~\cite{routerdc}} & \multicolumn{3}{c}{\cellcolor{Gray}WebRouter} \\ \cline{2-16}
                      & \#steps     & acc.(\%)   & price($\$$)   & \#steps     & acc.(\%)   & price($\$$) & \#steps     & acc.(\%)   & price($\$$) & \#steps     & acc.(\%)   & price($\$$) & \cellcolor{Gray}\#steps     & \cellcolor{Gray}acc.(\%)   & \cellcolor{Gray}price($\$$)   \\ \midrule
Apple        & 22.26   &34.8   &0.98    &11.39   &78.3   &0.52    &7.65    &87.0   &0.80    &9.13    &78.3   &0.16    &\cellcolor{Gray}9.00    &\cellcolor{Gray}78.3   & \cellcolor{Gray}0.12   \\
Arxiv        & 21.43   &34.8   &1.21    &13.04   &82.6   &0.84    &7.65    &91.3   &1.04    &10.00   &69.6   &0.24    &\cellcolor{Gray}8.70    &\cellcolor{Gray}87.0   &\cellcolor{Gray}0.14   \\
Coursera     & 21.22   &52.2   &1.03    &10.61   &82.6   &0.49    &8.52    &82.6   &1.04    &7.57    &78.3   &0.14    &\cellcolor{Gray}7.96    &\cellcolor{Gray}82.6   &\cellcolor{Gray}0.12   \\
Google       & 16.13   &65.2   &0.80    &7.52    &87.0   &0.34    &5.70    &82.6   &0.94    &9.00    &31.8   &0.17    &\cellcolor{Gray}6.14    &\cellcolor{Gray}82.6   &\cellcolor{Gray}0.08   \\
Huggingface  & 21.43   &34.8   &1.05    &8.52    &78.3   &0.36    &8.61    &87.0   &1.08    &11.30   &81.0   &0.21    &\cellcolor{Gray}10.10   &\cellcolor{Gray}81.0   &\cellcolor{Gray}0.16   \\ \midrule 
Average      & 20.49   &44.3   &1.01    &10.22   &81.7   &0.51    &\textbf{7.63}    &\textbf{86.1}   &0.98    &9.40    &67.8   &\underline{0.18}    &\cellcolor{Gray}\underline{8.38}    &\cellcolor{Gray}\underline{82.3}   &\cellcolor{Gray}\textbf{0.12}   \\ \bottomrule
\end{tabular}
}
\end{table*}
\section{Experiment}
\subsection{Experiment Setup}
\noindent\textbf{Dataset.}
We evaluate WebRouter on five real-world websites from WebVoyager dataset~\cite{Webvoyager}: Apple, Arxiv, Coursera, Google, and Huggingface. Following the data preprocessing in~\cite{browseruse}, each website contains at least 46 distinct tasks for evaluation. 
Due to the significant API costs, our training set, $\mathcal{D}_{\text{train}}$, is limited to 11,800 samples.

\noindent\textbf{Baselines.}
We compare WebRouter against strong baselines, including the state-of-the-art web agent, \emph{i.e.,} browser-use\cite{browseruse} configured with individual LLM backbones, and the RouterDC~\cite{routerdc} router. All models were accessed through the OpenRouter API, and all pricing is based on their listed rates: Gemini-2.5Flash ($0.30/M$ in, $2.50/M$ out), GPT-4.1Mini ($0.40/M$, $1.60/M$), and GPT-4o ($5/M$, $15/M$).

\begin{figure}[t]
    \centering
    \begin{subfigure}[t]{0.24\textwidth}
        \centering
        \includegraphics[width=\textwidth]{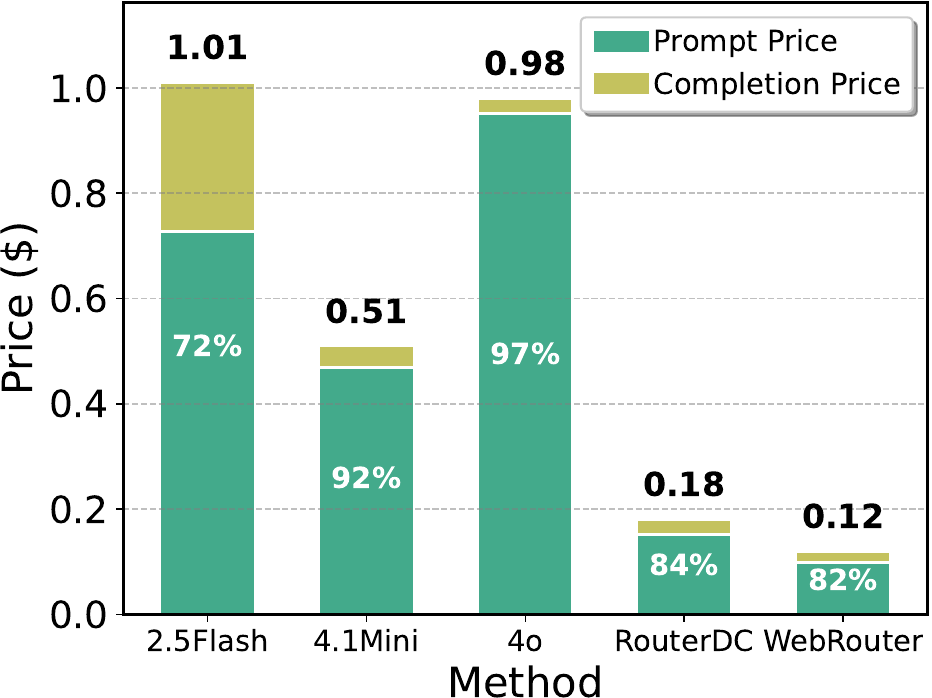}
        \caption{Price breakdown.}
    \end{subfigure}%
    ~ 
    \begin{subfigure}[t]{0.24\textwidth}
        \centering
        \includegraphics[width=\textwidth]{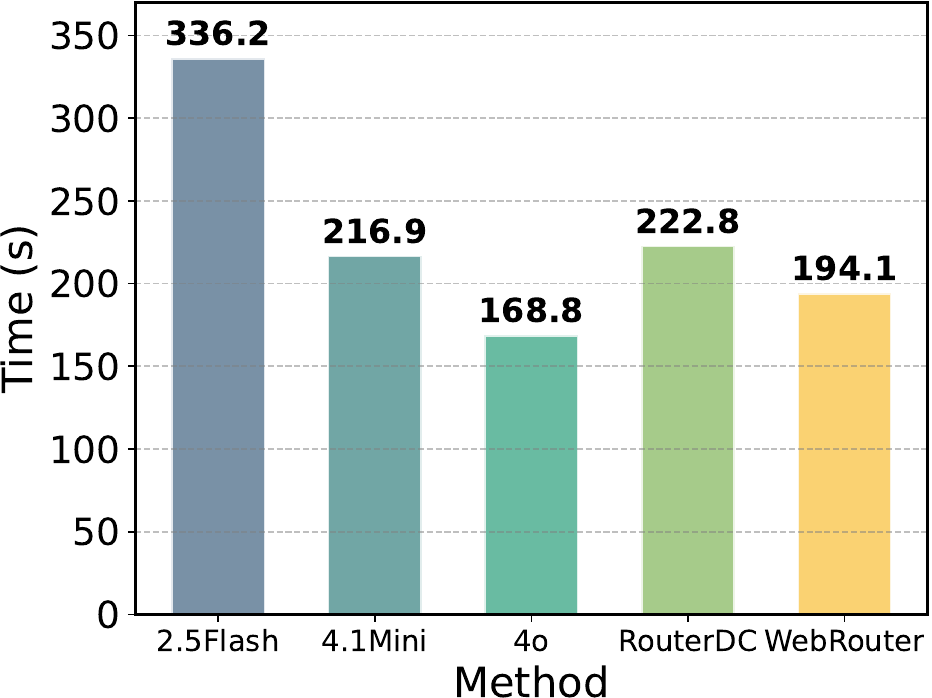}
        \caption{Average running time.}
    \end{subfigure}%
    \caption{Analysis of cost composition and execution time.}
    \label{fig:indepth}
\end{figure}

\noindent\textbf{Implementation Details.}
We use mDeBERTaV3-base~\cite{he2021deberta} as the query encoder $\psi$ with a 768-dimensional embedding space. The router is trained for 2000 steps using the AdamW optimizer with a learning rate of $2\times 10^{-5}$, which we selected based on its stable convergence as shown in Fig~\ref{fig:train}.
We set ca-VIB loss with $\beta=0.3$ and $\lambda=0.2$.

\subsection{Experimental Result}
\noindent\textbf{Main Results.}
Table~\ref{tab:main_res} presents our main results, evaluating performance across task accuracy (acc.), the average price, and the average number of steps. WebRouter demonstrates a superior performance-to-cost trade-off. This is most evident when compared to browser-use with GPT-4o: WebRouter reduces the operational cost by a striking 87.8\% (from 0.98\$ to 0.12\$) while incurring only a minor 3.8\% drop in accuracy. 
This cost reduction is a direct outcome of the cost-regularization term in our ca-VIB loss (Eq.~\ref{eq:ca_vib_loss}), which explicitly trains the router to favor less expensive models. 
Crucially, this efficiency is achieved while maintaining high accuracy because the VIB successfully distills the true complexity from verbose prompts. Furthermore, WebRouter outperforms the competing routing baseline, RouterDC, in both accuracy (82.3\% vs. 67.8\%) and efficiency (8.38 vs. 9.40 steps), which we attribute to VIB's superior ability to handle the informational redundancy inherent in web agent queries.


\noindent\textbf{Cost and Running Time.}
Fig.~\ref{fig:indepth}(a) analyzes the cost composition, revealing that prompt tokens, as defined in Eq.~\eqref{eq:cost}, contribute to over 70\% of the total price for all models. This highlights the importance of our cost-aware ca-VIB loss, which directly penalizes the selection of inherently expensive models at routing time. In terms of running time(Fig.~\ref{fig:indepth}(b)), WebRouter is only marginally slower than the GPT-4o baseline (14\%), demonstrating that its significant 87.8\% price reduction comes at a minimal cost to execution speed.

\begin{figure}[t]
    \centering
    \begin{subfigure}[t]{0.30\textwidth}
        \centering
        \includegraphics[width=\textwidth]{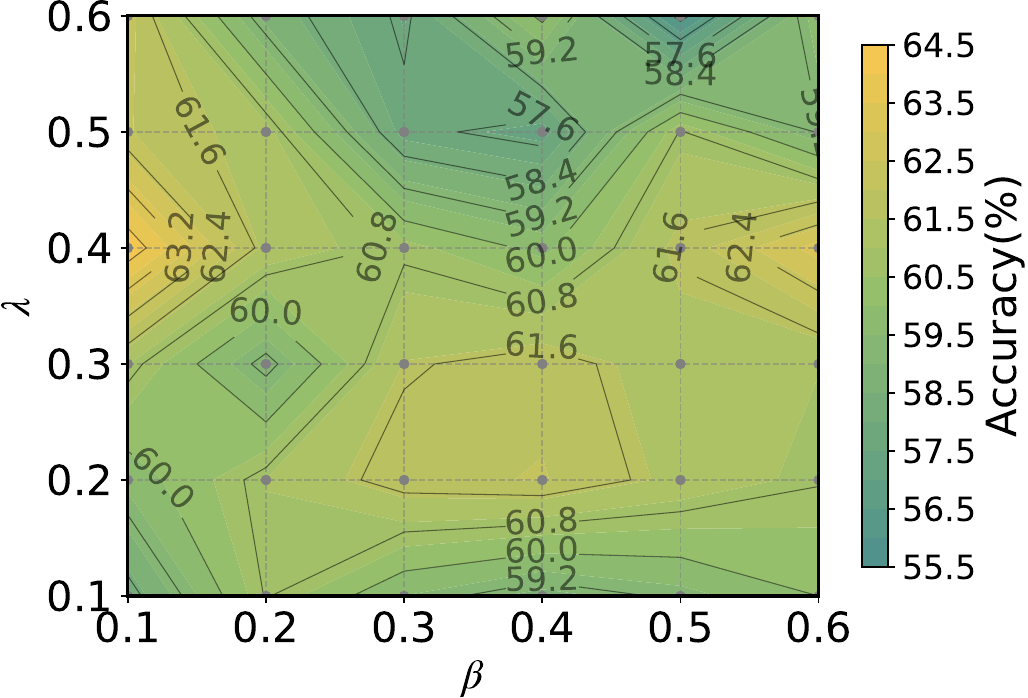}
        \caption{Hyperparameter sensitivity.}
    \end{subfigure}%
    ~ 
    \begin{subfigure}[t]{0.20\textwidth}
        \centering
        \includegraphics[width=\textwidth]{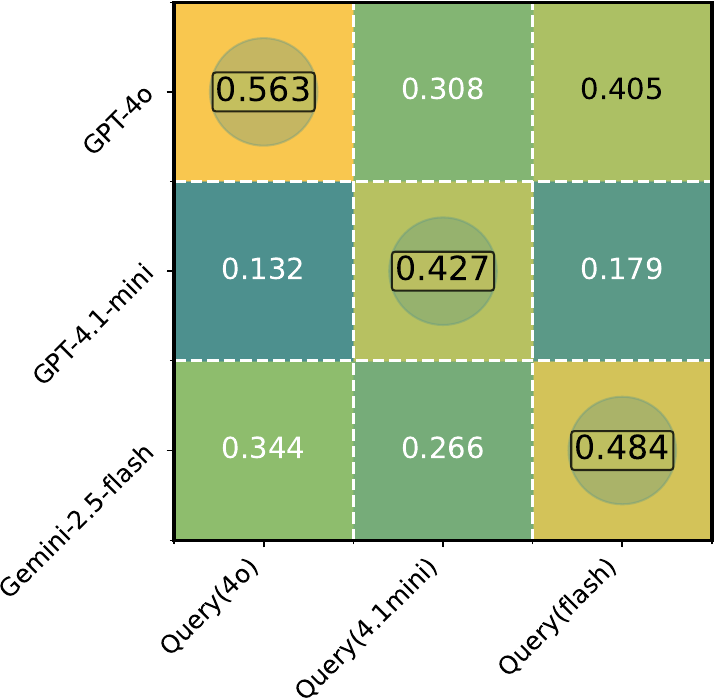}
        \caption{Learned query affinity.}
    \end{subfigure}
    \caption{Analysis of hyperparameters and representations.}
    \label{fig:analysis}
\end{figure}

\noindent\textbf{Ablation and Analysis.}
We analyze the sensitivity of our key hyperparameters $\beta$ and $\lambda$ in Fig.~\ref{fig:analysis}(a). The model's performance is stable across a range of values, with optimal accuracy achieved around $\lambda=0.4$. 
Fig.~\ref{fig:analysis}(b) visualizes the router's learned representations, showing a clear affinity between queries and the embeddings of their optimally selected LLMs (measured by cosine similarity). This indicates that our VIB-based approach successfully learns meaningful, task-relevant features for the routing decision.

\section{Conclusion}
In this paper, we introduced WebRouter, a novel query-specific router for LLM-brained web agents. 
We addressed the dual challenges of high operational costs and noisy, redundant input prompts by proposing a cost-aware VIB loss. 
Our experiments show that this information-theoretic approach greatly reduces web agent operating costs with minimal impact on task accuracy.

\vfill\pagebreak

\bibliographystyle{IEEEbib}
\bibliography{refs}

\end{document}